\documentclass[runningheads,a4paper]{llncs}

\usepackage{lineno,hyperref}
\usepackage{amssymb}
\usepackage{graphicx}
\usepackage{amsmath,amssymb,amscd}
\usepackage{algorithm}
\usepackage{algpseudocode}
\usepackage{subfigure}
\usepackage{multirow}

\usepackage{url}

\begin{document}

\mainmatter  % start of an individual contribution

\title{Dictionary Learning and Sparse Coding-based Denoising for High-Resolution Task Functional Connectivity MRI Analysis}

\titlerunning{DLSC-based Denoising for High-Resolution tfMRI Analysis}

\author{Seongah Jeong$^{\dagger}$\and Xiang Li$^{\ddagger}$\and Jiarui Yang$^{\ddagger}$\and Quanzheng Li$^{\ddagger}$\and\\Vahid Tarokh$^{\dagger}$}
\authorrunning{Seongah Jeong et al.}

\institute{$^\dagger$ School of Engineering and Applied Sciences (SEAS), \\ Harvard University, 29 Oxford street, Cambridge, MA 02138, USA\\ $^\ddagger$ Department of Radiology, Massachusetts General Hospital and Harvard Medical School, Boston, MA 02114, USA}


\toctitle{Lecture Notes in Computer Science}
\tocauthor{Authors' Instructions}
\maketitle

\begin{abstract}
We propose a novel denoising framework for task functional Magnetic Resonance Imaging (tfMRI) data to delineate the high-resolution spatial pattern of the brain functional connectivity via dictionary learning and sparse coding (DLSC). In order to address the limitations of the unsupervised DLSC-based fMRI studies, we utilize the prior knowledge of task paradigm in the learning step to train a data-driven dictionary and to model the sparse representation. We apply the proposed DLSC-based method to Human Connectome Project (HCP) motor tfMRI dataset. Studies on the functional connectivity of cerebrocerebellar circuits in somatomotor networks show that the DLSC-based denoising framework can significantly improve the prominent connectivity patterns, in comparison to the temporal non-local means (tNLM)-based denoising method as well as the case without denoising, which is consistent and neuroscientifically meaningful within motor area. The promising results show that the proposed method can provide an important foundation for the high-resolution functional connectivity analysis, and provide a better approach for fMRI preprocessing.
\keywords{Functional Magnetic Resonance Imaging (fMRI), Denoising, Dictionary Learning, Sparse Coding, Connectivity}
\end{abstract}

%%%%%%%%%%%%%%%%%%%%%%%%%%%%%%%%%%%%%%%%%%%%%%%%%%%%%%%%%%%%%%%%%%%%%%%%%%%%%%
%% Section 1
\section{Introduction}
Functional magnetic resonance imaging (fMRI) enables the inference of functional connectivity among different brain regions \cite{Biswal95MRM,Fox07Nature,Yeo11JN,Buckner11JN,Jeong17OHBM} and the early diagnosis of various brain disorders \cite{Wee12NI,Jie14TBME}. However, the functional connectivity analysis based on the fMRI data is limited in accuracy and reliability due to the inherently low signal-to-noise ratio (SNR) of fMRI signals which results from the high intrinsic noise and small magnitude of Blood Oxygenation Level Dependent (BOLD) signals. To address this issue, the preprocessing to reduce the noise and improve the SNR is typically carried out either by improving the sample size (i.e. group-wise study with multiple subjects), or via filtering \cite{Biswal95MRM,Fox07Nature,Yeo11JN,Buckner11JN,Wink04TMI,Bhushan16Plos}, e.g., Gaussian low pass (GLP)-based filter \cite{Wink04TMI} and non-local means (NLM)-based filter \cite{Bhushan16Plos}.

In this work, we develop a novel denoising framework for task fMRI (tfMRI) data to enable the high-resolution functional connectivity analysis via dictionary learning and sparse coding (DLSC)-based method. The proposed method exploits the sparsity of the underlying fMRI signals, by which the temporal dynamics at each voxel can be represented as a sparse combination of basis functions to characterize global brain dynamics \cite{Lv15Media,Lv15TBME}. In order to overcome the limitations of the current unsupervised DLSC-based fMRI studies such as absence of the interpretation of the output and loss of statistical power, we utilize the prior knowledge of task paradigm in the learning step to train a data-driven dictionary and to model the sparse representation \cite{Zhao15TMI}. We apply the proposed DLSC-based denoising method to the tfMRI data acquired during motor task from Human Connectome Project (HCP) \cite{Barch13NI}. It is observed that the proposed denoising method can exert more pronounced effects on the connectivities with high correlation, but less pronounced effects on the connectivities with low correlation which are estimated from the raw signals, in comparison to the baseline method. This allows us to obtain a more distinct spatial connectivity pattern and to achieve the high-resolution functional connectivity analysis.     

%%%%%%%%%%%%%%%%%%%%%%%%%%%%%%%%%%%%%%%%%%%%%%%%%%%%%%%%%%%%%%%%%%%%%%%%%%%%%%
%% Section 2
\section{Methods}
In this section, we first briefly describe the HCP motor tfMRI dataset \cite{Barch13NI}, and then provide the detailed description of the proposed denoising framework.    

\begin{figure}[t]
\begin{center}
\includegraphics[width=10cm]{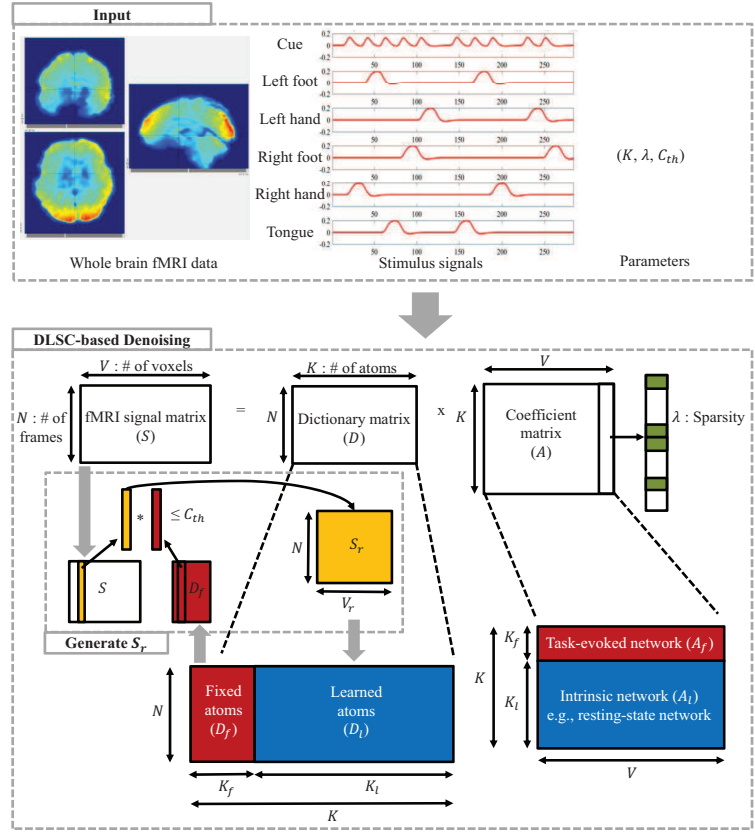}
\caption{The framework of the DLSC-based denoising method.} \label{fig:frame}
\end{center}
\end{figure}
%%%%%%%%%%%%%%%%%%%%%%%%%%%%%%%%%%%%%%%%%%%%%%%%%%%%%%%%%%%%%%%%%%%%%%%%%%%%%%
%% Section 2-1
\subsection{Dataset}\label{sec:data}
In this work, we use the minimally preprocessed fMRI data (TR = $720$ ms, TE = $33.1$ ms, $2 \times 2 \times 2$ mm voxel) from 50 subjects provided in HCP Q1 release \cite{Barch13NI}. The details for data acquisition and experiment design can be found in \cite{Barch13NI}. The tfMRI data are obtained during motor task, where the participants are informed with visual cues to tap their left or right fingers, squeeze their left or right toes, and move their tongue. This task is verified to be able to identify the effector corresponding to the specific activation individually in \cite{Yeo11JN,Buckner11JN}. Each run consists of $13$ blocks with $2$ tongue movements, $2$ right and $2$ left hand movements, $2$ right and $2$ left foot movements and three $15$ s fixation blocks, where each block lasts $12$s and is preceded by a $3$s visual cue. The number of frames or time samples for each subject is $284$ and the total run duration is $214$ s.

%%%%%%%%%%%%%%%%%%%%%%%%%%%%%%%%%%%%%%%%%%%%%%%%%%%%%%%%%%%%%%%%%%%%%%%%%%%%%
%% Section 2-2
\subsection{Dictionary Learning and Sparse Coding (DLSC)-based Denoising} 
The proposed DLSC-based denoising method aims at representing the fMRI signal at each voxel as a sparse linear combination of dictionary basis functions or atoms characterizing global brain dynamics. Unlike the existing unsupervised DLSC-based fMRI studies, we utilize the prior knowledge of task paradigm in the learning step to train the data-driven dictionary and to develop the sparse coding as illustrated in Fig. \ref{fig:frame}. 

Specifically, for each subject, we factorize the $N \times V$ fMRI signal matrix $S$, with $N$ being the number of frames and $V$ being the number of voxels, into $S = DA$, where $D$ is the $N \times K$ dictionary matrix with $K$ being the number of atoms and $A$ is the corresponding $K \times V$ coefficient matrix. In the proposed DLSC-based scheme, we consider two different sets of dictionary atoms such as fixed atoms and learned atoms. Accordingly, the dictionary matrix $D$ is composed as $D = [D_f\,\,D_l]$, where the sub-dictionary matrices $D_f$ and $D_l$ consist of the $K_f$ fixed atoms and the $K_l$ learned atoms, respectively, which correspond to the sub-coefficient matrices $A_f$ and $A_l$ constructing the coefficient matrix $A=[A_f; \,A_l]$. The fixed atoms are predefined as the task stimulus curves which are generated by the convolution of a Statistical Parametric Mapping (SPM) \cite{SPM12} canonical hemodynamic response function (HRF) and the simple boxcar stimulus function indicating each occurrence of a generation event. Here, we consider the six different stimulus curves for the visual cues, left hand, left foot, right hand, right foot and tongue movements in Fig. \ref{fig:frame}. The learned atoms are trained in unsupervised way with the reconstructed signal matrix $S_r$ by solving the following minimization problem:
\begin{subequations}\label{eq:dl}
\begin{eqnarray}
&& \hspace{-1.5cm} \underset{D_l, A^*}{\text{minimize}} \hspace{0.5cm} \left\|S_r-D_lA^*\right\|_F^2 \\
&& \hspace{-1cm}  \text{s.t.} \hspace{0.9cm} \left\|a^*_v\right\|_0 \le \lambda, \hspace{0.2cm} v = 1, \dots, V_r,
\end{eqnarray}
\end{subequations}  
where $\|\cdot\|_F$ and $\|\cdot\|_0$ indicate the Frobenius norm and the zero norm, respectively; $S_r$ is the $N \times V_r$ signal matrix reconstructed from the original signal matrix $S$ which only includes the voxels with the correlation value with the fixed atoms less than predefined threshold $C_{th}$ to satisfy the condition $V_r \ge K_l$; $A^* = [a^*_1 \cdots a^*_{V_r}]$ is the $K_l \times V_r$ coefficient matrix; and $\lambda$ represents the sparsity constraint on the maximum number of non-zero coefficients for signal at each voxel of the matrix $S_r$. It is noted that the reconstructed signal matrix $S_r$ is considered for mitigating the malfunction caused by the intra-correlation between the fixed dictionary $D_f$ and the learned dictionary $D_l$. In order to solve the minimization problem (\ref{eq:dl}), K-SVD algorithm \cite{Aharon06TSP} and orthogonal matching pursuit (OMP) \cite{Pati093Asi} are adopted, where K-SVD algorithm coupled with OMP optimizes the dictionary $D_l$ in an iterative and alternative fashion with the sparse coding $A^*$. Then, the coefficient matrix $A$ for the original signal matrix $S$ is determined with the final dictionary matrix $D$ generated from the sub-dictionaries $D_f$ and $D_l$ by using OMP. The proposed DLSC-based denoising framework can be readily applied to the other tfMRI datasets, e.g., emotion, gambling, language, relational, social and working memory tasks \cite{Barch13NI}, but also used with other functional neuroimaging methods.    

For the DLSC-based denoising method, the parameter tuning for the dictionary size $K$, the sparsity $\lambda$ and the threshold $C_{th}$ is required which affects the denoising performance. However, since there is no gold criterion for the choice of these parameters, we perform a grid-search for finding the best parameter combinations with $K=300$ to $500$, with step of $100$, $\lambda = 5$ to $50$, with step of $5$, and $C_{th}=0.1$ to $0.4$, with step of $0.1$, and finally choose $(K, \lambda, C_{th}) = (400, 40, 0.1)$ based on visual inspection of resulting connectivity performance. 
  
%%%%%%%%%%%%%%%%%%%%%%%%%%%%%%%%%%%%%%%%%%%%%%%%%%%%%%%%%%%%%%%%%%%%%%%%%%%%%%
%% Section 3
\section{Results}\label{sec:connectivity}
\begin{figure}[t]
\begin{center}
\includegraphics[width=7cm]{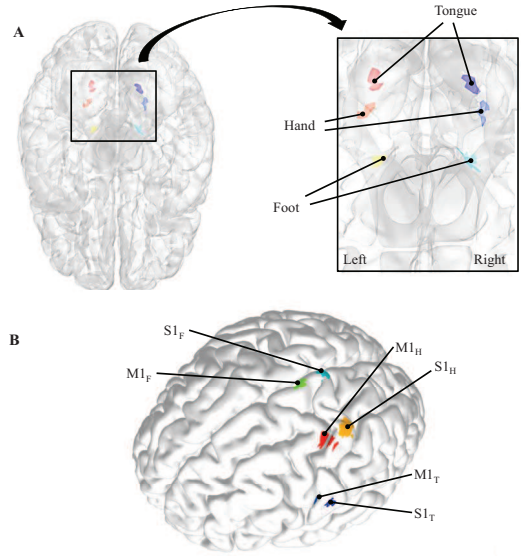}
\caption{Cerebrocerebellar circuits involved in somatomotor networks. A: Cerebellar seed regions corresponding to foot, hand and tongue tasks. B: M$1_\text{F}$, S$1_\text{F}$, M$1_\text{H}$, S$1_\text{H}$, M$1_\text{T}$ and S$1_\text{T}$ in the left cerebral cortex.} \label{fig:ROI}
\end{center}
\end{figure}

In this work, we focus on the cerebrocerebellar circuits involved in somatomotor networks to analyze the validity of proposed method. To this end, the ground truth functional connectivity as reported in \cite{Yeo11JN,Buckner11JN} between three cerebellar seed regions for the foot (F), hand (H) and tongue (T) tasks and six cerebral regions (M$1_\text{F}$, S$1_\text{F}$, M$1_\text{H}$, S$1_\text{H}$, M$1_\text{T}$ and S$1_\text{T}$) related to the corresponding seed regions in the left and right hemisphere are considered (see, Fig. \ref{fig:ROI}). Each region consists of a single surface vertex ($4 \times 4$ mm) centering at the MNI coordinate tabulated in \cite{Yeo11JN,Buckner11JN}, and the visualization is performed via BrainNet \cite{Xia13Plos}. For comparison, the temporal non-local mean (tNLM)-based denoising method \cite{Bhushan16Plos} is adopted which can achieve significant performance improvement in comparison to the linear filter-based denoising. The tNLM-based method effectively substitutes the spatial similarity weighting in standard NLM with a weighting that is based on the correlation between time series. For the tNLM-based denoising method, we set the distance parameter $= 11$ and the smoothing level $= 0.72$ following \cite{Bhushan16Plos} to empirically provide good results with reasonable computational cost.        

\begin{figure*}[h!]
\centering
\subfigure{\label{fig:F}\includegraphics[width=6.1cm]{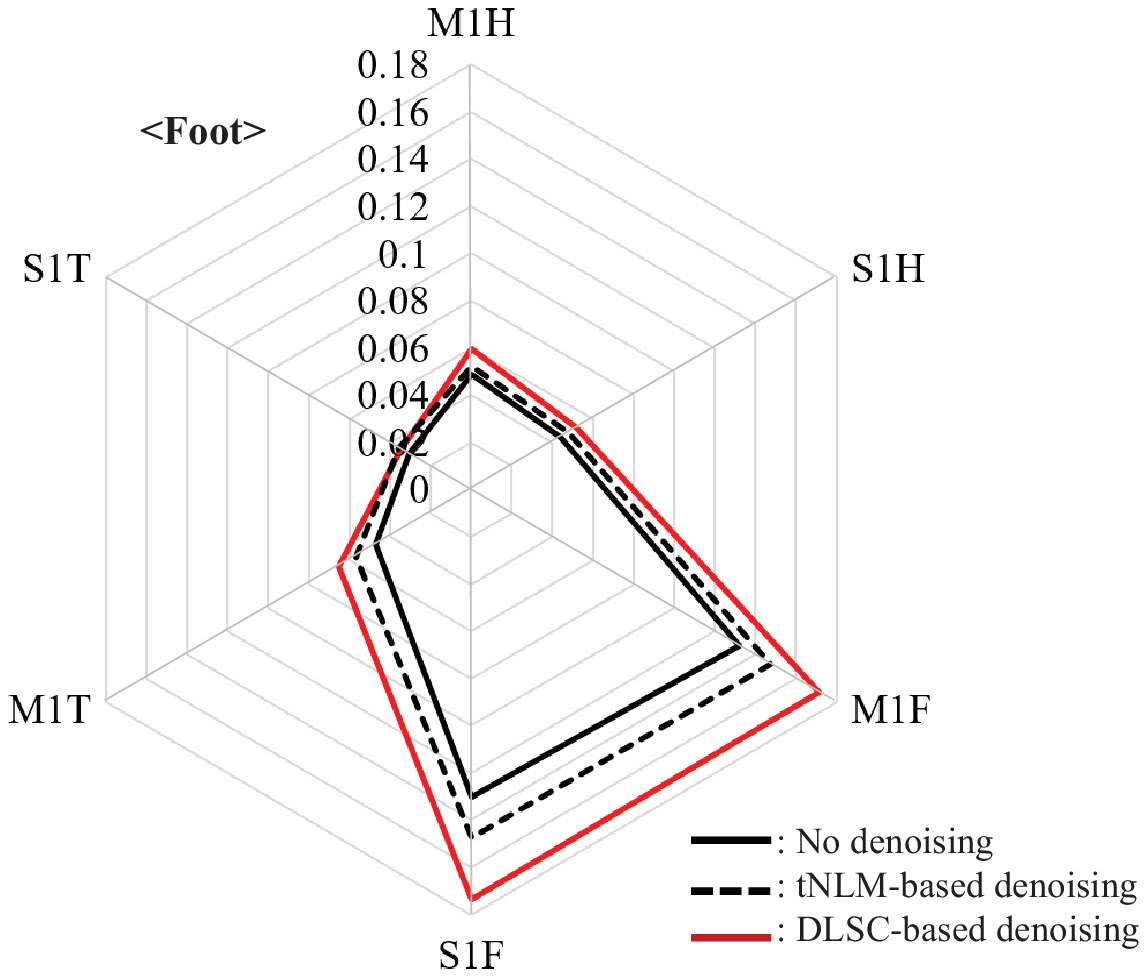}}\\
\subfigure{\label{fig:H}\includegraphics[width=6.1cm]{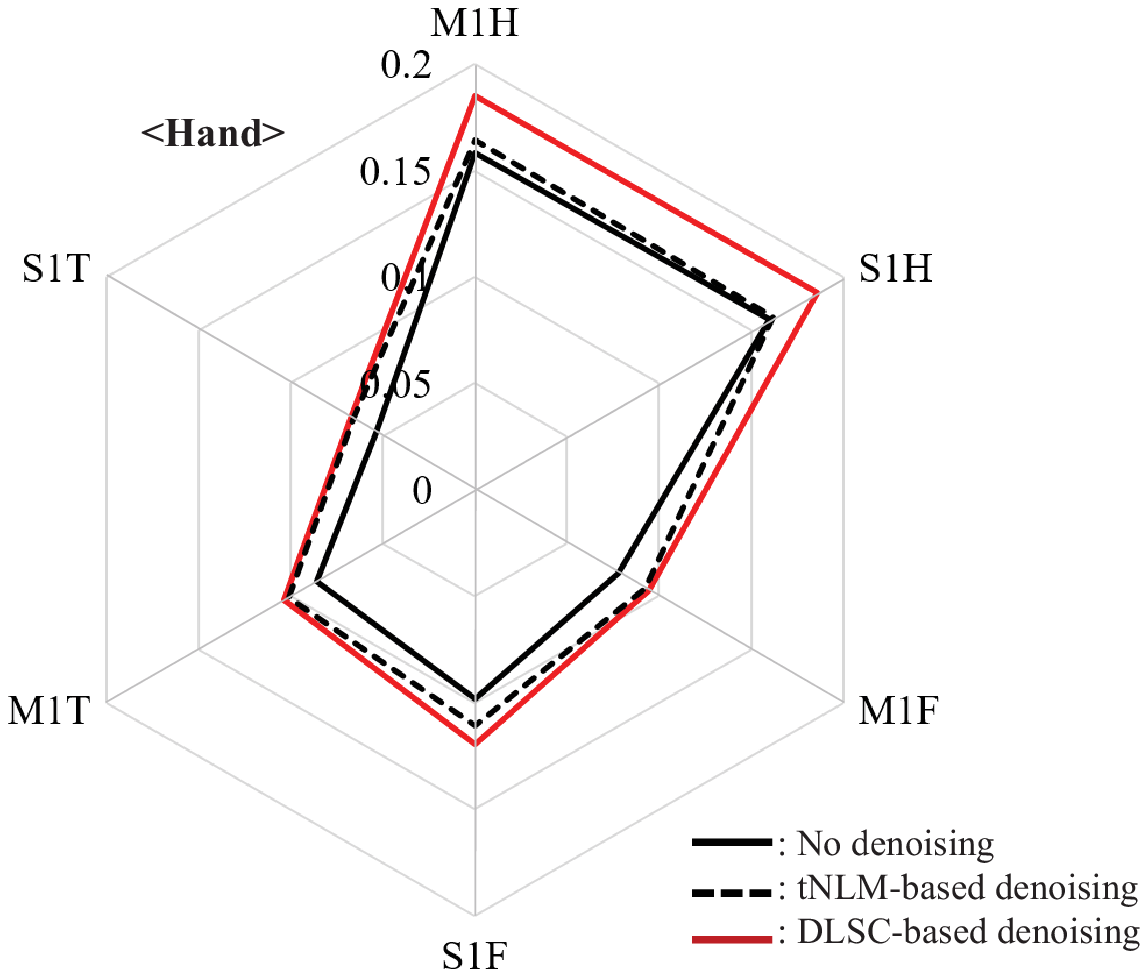}}\\
\subfigure{\label{fig:T}\includegraphics[width=6.1cm]{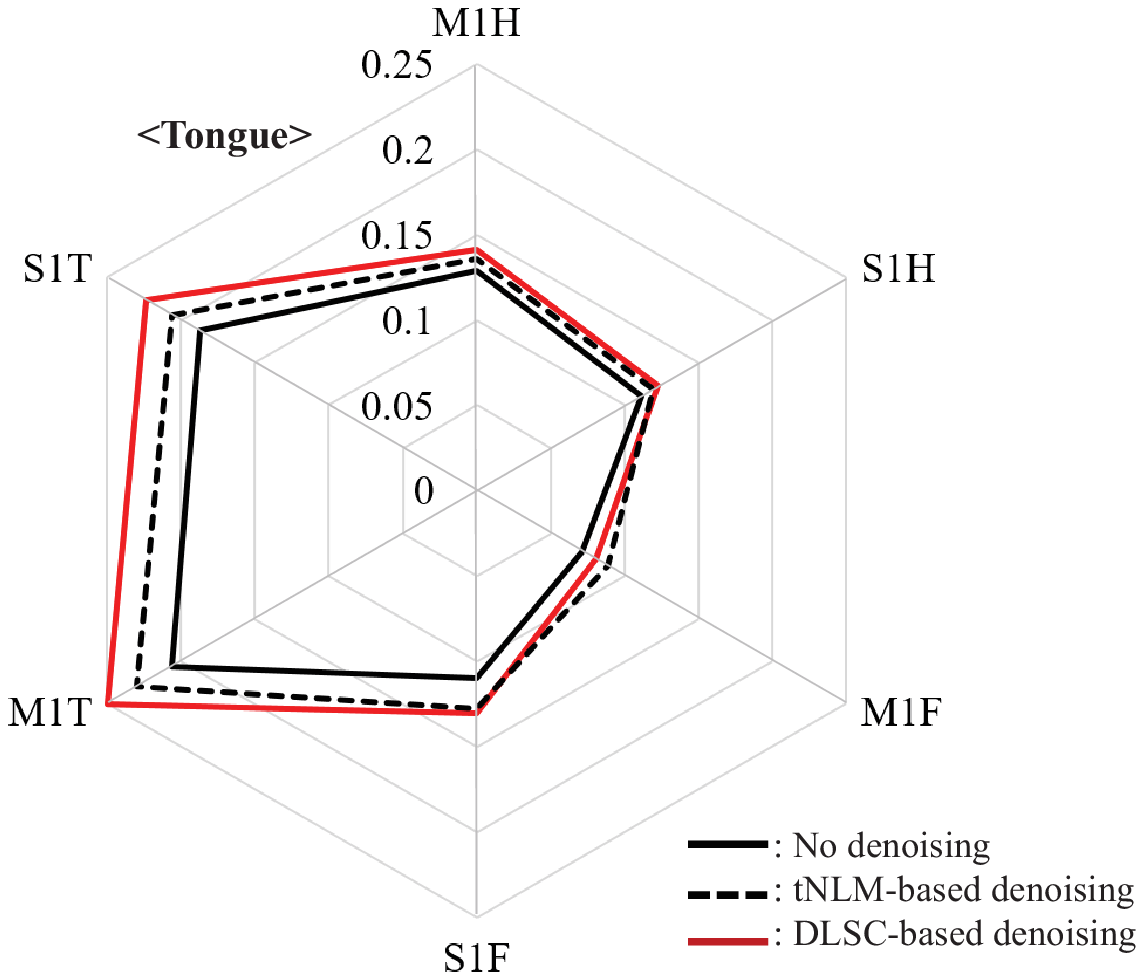}} \\
\caption{Quantitative measures of functional connectivity strength of unprocessed data (black solid line), tNLM-based denoised data (black dashed line) and DLSC-based denoised data (red solid line) for cerebellar anterior lobe seed regions linked to the foot, hand and tongue representations}
\label{fig:connectivity}
\end{figure*}  

The group-averaged functional connectivity maps for the raw data (black solid line) and the denoised data by using tNLM-based method (black dashed line) and our proposed DLSC-based method (red solid line) are visualized for the foot, hand and tongue tasks in Fig. \ref{fig:connectivity}. The functional connectivity is evaluated by computing the Pearson's correlation coefficient between contralateral cerebral and cerebellar regions and averaged across the hemispheres. Fisher's r-to-z transformation and its inverse transformation are further applied to promote the normality of the distribution of correlations. 

In Fig. \ref{fig:connectivity}, it is observed that our novel denoising framework can strengthen the connectivity between the seed regions associated with each particular movement and the corresponding M1 and S1 regions, that is, the connectivities at M$1_\text{F}$ and S$1_\text{F}$ for foot task, M$1_\text{H}$ and S$1_\text{H}$ for hand task, and M$1_\text{T}$ and S$1_\text{T}$ for tongue task can be strikingly pronounced by the proposed DLSC-based denoising method. Furthermore, the emphasis effect of the DLSC-based method is more significant on the connectivities with high correlation, but less on the connectivities with low correlation estimated from the raw signals, which is consistent across the tasks and subjects. This can provide the neuroscientifically meaningful within motor area to enable the high-resolution functional connectivity analysis, which cannot be attained by the tNLM-based denoising method nor with no denoising method applied. For example, the highest connectivity emphasis effect of the tNLM-based method is on M$1_\text{T}$ and M$1_\text{F}$ in the hand movement of Fig. \ref{fig:connectivity}. In addition, the DLSC-based method shows the strongest connectivity patterns for all motor tasks than the tNLM-based method. We believe that, from these important observations, the proposed DLSC-based denoising method can effectively recover the disrupted functional connectivity by artifacts and noise, and therefore have the potential to discover the previously hidden spatial connectivity pattern. 

%%%%%%%%%%%%%%%%%%%%%%%%%%%%%%%%%%%%%%%%%%%%%%%%%%%%%%%%%%%%%%%%%%%%%%%%%%%%%%
%% Section 4
\section{Discussion}
In this work, we have studied a DLSC-based denoising framework for the high-resolution tfMRI functional connectivity analysis by using the sparseness of the underlying hemodynamic signals in brain. Unlike the traditional unsupervised DLSC-based fMRI studies, we utilize the prior knowledge of task paradigm in the learning step to train the dictionary and to develop the sparse coding. The denoising effect of the proposed framework is evaluated by applying it to the publicly available HCP motor tfMRI dataset. Studies on the functional connectivity between cerebellar seed regions and cerebral regions in somatomotor networks show the significant denoising performance of the proposed scheme, in comparison to tNLM-based denoising method as baseline scheme as well as no denoising case. It is observed that the DLSC-based denoising method can exert more pronounced effects on the connectivities with high correlation, but less pronounced effects on the connectivities with low correlation estimated from the raw signals, which is consistent and neuroscientifically meaningful within motor area. This shows the capability of the proposed DLSC-based denoising framework to provide a more distinct spatial pattern and accordingly to enable the high-resolution functional connectivity analysis.

\end{document}